\documentclass[sigconf,authorversion]{acmart}

\usepackage{color}
\usepackage{float}
\usepackage{graphicx}
\usepackage{listings}
\usepackage{url}
\usepackage{subcaption}
\usepackage{algorithm}
\usepackage{tabularx}
\usepackage{color}
\usepackage{balance}
\usepackage{booktabs} 

\usepackage{pgfplotstable}

\definecolor{codegreen}{rgb}{0,0.6,0}
\definecolor{codegray}{rgb}{0.5,0.5,0.5}
\definecolor{codepurple}{rgb}{0.58,0,0.82}
\definecolor{backcolour}{rgb}{255,255,255}

\lstdefinestyle{mystyle}{
  backgroundcolor=\color{backcolour},
  commentstyle=\color{codegreen},
  keywordstyle=\color{magenta},
  numberstyle=\tiny\color{codegray},
  stringstyle=\color{codepurple},
  basicstyle=\footnotesize,
  breakatwhitespace=false,         
  breaklines=true,                 
  captionpos=b,                    
  keepspaces=true,                 
  numbers=left,                    
  numbersep=5pt,                  
  showspaces=false,                
  showstringspaces=false,
  showtabs=false,                  
  tabsize=2
}

\setcitestyle{nosort}

\copyrightyear{2023} 
\acmYear{2023} 
\setcopyright{rightsretained} 
\acmConference[DESTION 2023]{Design Automation for CPS and IoT}{May 9--12, 2023}{San Antonio, TX, United States}

\begin{document}

\title[]{Constrained Bayesian Optimization for Automatic Underwater Vehicle Hull Design}

\author{Harsh Vardhan}
\email{harsh.vardhan@vanderbilt.edu}
\affiliation{%
  \institution{Vanderbilt University}
  \city{Nashville}
  \state{TN}
  \country{USA}
}

\author{Peter Volgyesi}
\email{peter.volgyesi@vanderbilt.edu}
\affiliation{%
  \institution{Vanderbilt University}
  \city{Nashville}
  \state{TN}
  \country{USA}
}

\author{Will Hedgecock}
\email{ronald.w.hedgecock@vanderbilt.edu}
\affiliation{%
  \institution{Vanderbilt University}
  \city{Nashville}
  \state{TN}
  \country{USA}
}

\author{Janos Sztipanovits}
\email{janos.sztipanovits@vanderbilt.edu}
\affiliation{%
  \institution{Vanderbilt University}
  \city{Nashville}
  \state{TN}
  \country{USA}
}

\renewcommand{\shortauthors}{Authors et al.}

\begin{abstract}
Automatic underwater vehicle hull Design optimization is a complex engineering process for generating a UUV hull with optimized properties on a given requirement. First, it involves the integration of involved computationally complex engineering simulation tools.  Second, it needs integration of a sample efficient optimization framework with the integrated toolchain.  To this end, we integrated the CAD tool called FreeCAD with CFD tool openFoam for automatic design evaluation. For optimization, we chose Bayesian optimization (BO), which is a well-known technique developed for optimizing time-consuming expensive engineering simulations and has proven to be very sample efficient in a variety of problems, including hyper-parameter tuning and experimental design. During the optimization process, we can handle infeasible design as constraints integrated into the optimization process. By integrating domain-specific toolchain with AI-based optimization, we executed the automatic design optimization of underwater vehicle hull design. For empirical evaluation, we took two different use cases of real-world underwater vehicle design to validate the execution of our tool. 

\end{abstract}


\maketitle

\section{Introduction}
\label{sec:intro}
Artificial Intelligence (AI) and Machine Learning (ML) are becoming increasingly useful for both system-level design~\cite{allard2014unmanned,vardhan2021machine,alam2015design,vardhan2022deep} and control~\cite{vardhan2021rare,abbeel2010autonomous,vardhan2022reduced}. However, its application to real-world designs is still nascent for a variety of reasons. The first challenge is that these areas involve complex tools and require cumbersome, integrated toolchains. Integrating these tools requires domain knowledge, as well as the development of tool-level integration platforms. The next challenge involves the complexity of carrying out evaluations due to the high cost of data labeling. In this work, we aimed to address both challenges in the context of an underwater vehicle hull design optimization problem. The design of an underwater vehicle hull involves multiple steps: CAD design of the hull, generation of stereolithography (STL) files, generation of a computational fluid dynamics (CFD) simulation environment, creation of volume meshing, fixing of the initial and boundary conditions, and solution of Navier-Stokes equations at the mesh level. For a fully automated design optimization process, it is necessary to integrate the execution of all of these processes in a single tool.

Another aspect of this work relates to optimization algorithms. There are many gradient-free and gradient-based algorithms available; however, in recent times, Bayesian Optimization (BO)~\cite{clark1961greatest,movckus1975bayesian,zhilinskas1975single} has emerged as a well-established paradigm for optimizing expensive-to-evaluate functions in a sample-efficient manner, and it has been successfully applied to many scientific domains. Bayesian Optimization is a complex optimization process that provides the benefits of both model-based and simulation-based optimization approaches. To this end, BO creates a probabilistic model of an unknown function during the optimization process and uses this model to simulate and search for the best candidate sample to be used in the next evaluation. Constraints may exist which make some of the designs and design space infeasible, and these must be handled appropriately during the optimization process. 
In this work, we have made the following contributions:
\begin{enumerate}
    \item A ready-to-use underwater vehicle design tool using a Myring hull-based parametric CAD seed design,
    \item Integration of a constrained Bayesian Optimization framework for UUV hull design problem.
   
\end{enumerate}
 This fully automated optimization toolchain can be used as a good starting point to study and test different optimization methods on real-world complex problems, in addition to providing a useful tool for underwater vehicle hull designers. The code for running the experimentation and installing the toolchain can be found here \textcolor{blue}{\url{https://github.com/vardhah/ConstraintBOUUVHullDesign}}.

\section{Problem Formulation and Approach}
\label{sec:problem}
In this section, we formulate the UUV hull design problem as a constrained optimization problem. The hull shape of an underwater vehicle is indicated as $\Omega$ and can be defined using a multivariate parameter $x$, that is, $x \triangleq \Omega$. If $f$ is the objective function that maps a 3D shape $\Omega$ with a complex property of coupled two-way solid-fluid dynamics, that is, drag force ($F_d$) ($f:\Omega \mapsto F_d $.), then the optimization problem can be formulated as:
\begin{equation}
     \Omega^{*} =\underset{x \in DS }{\mathrm{argmin}} f(x)  
\end{equation}
Here, $DS$ is our design search space. The UUV hull contains electronics, sensors, and other mechanical and electrical components. Packing them into the hull imposes a non-linear constraint on the optimization process. The hull design problem can then be formulated as a constrained optimization problem defined as follows:
 \begin{gather}
     \Omega^{*} =\underset{x \in DS }{\mathrm{argmin}} f(X) \\
     s.t.\; g(x) \leq 0
\end{gather}
Here, constraint function $g(x)$ ensures that all selected components can be packed inside the designed UUV hull.  To solve this optimization problem, we utilize a constrained Bayesian Optimization framework as formulated by \cite{gardner2014bayesian}. 

\subsection{Constrained Bayesian Optimization}
\begin{figure}[ht!]
        \centering
        \captionsetup{justification=centering}
        \includegraphics[width=0.45\textwidth]{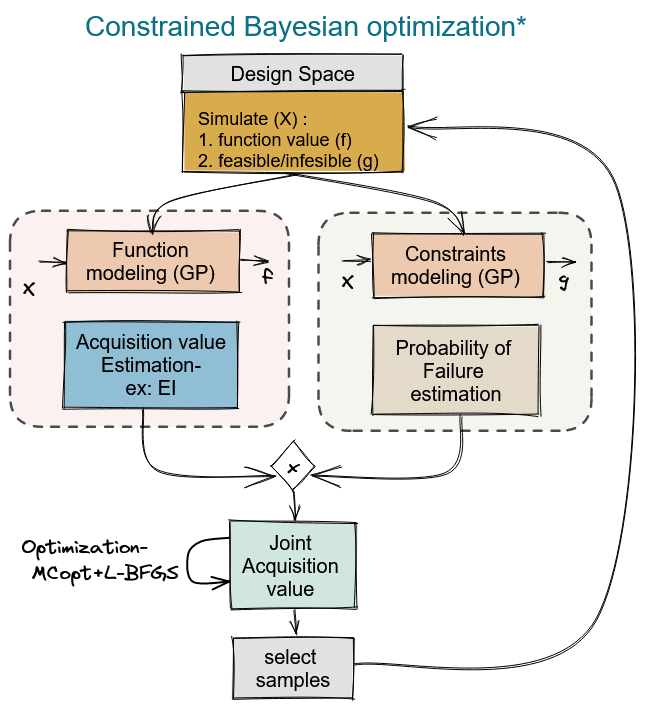}
        \caption{Constrained Bayesian Optimization - Overview}
        \label{fig:cbo}
    \end{figure}

Bayesian Optimization relies on a probabilistic model of the system of interest during optimization, and the fidelity of the model is the most decisive factor in the optimization process. We use the Gaussian process \cite{rasmussen2003gaussian} defined below to model system behavior ($f$):
\begin{equation}
     f\sim \mathcal{GP}(\mu(.),\kappa(.,.))\\\\
\end{equation}
Here $\mu(.)$ is the mean function and $\kappa(.,.)$ is the covariance kernel. For any given pair of input points $x,x' \in R^d$, these are defined as:
\begin{gather}
\mu(x)= \mathbb{E}[f(x)]\\
\kappa(x,x')= \mathbb{E}[(f(x)-\mu(x))(f(x')-\mu(x')]
\end{gather}
In the Bayesian sequential design optimization process, a crucial step at each iteration is to select the most promising candidate $x^*$ for evaluation in the next iteration. In the BO setting, this is done by defining an acquisition function. The design of an acquisition function is a critical component in the performance efficiency of the BO. Let $x^+$ be the best-evaluated sample so far. To select a candidate point $\hat x$ in the next iteration, an improvement is defined according to Mockus et al.~\cite{movckus1975bayesian} as follows:
\begin{equation}
    I(\hat{x})=max\{ 0,f(\hat{x})-f(x^+)\}
\end{equation}
The expected improvement in such a case is defined as an EI acquisition function, which has a closed-form solution for estimating it from a new candidate point, as given by Mockus et al.~\cite{movckus1975bayesian} and Jones et al.~\cite{jones1998efficient}: 
\begin{gather}
    EI(\hat x)= \mathbf{E}\big[ I(\hat{x})|\hat{x})| \big] \\
    EI(x^+)= (f(x^*)- \mu^+) \Phi(\frac{f(x^*)-\mu^+) }{\sigma^+}+ \sigma^+ \phi(\frac{f(x^*)-\mu^+)}{\sigma^+})
\end{gather}
Here, $\phi$ is the standard normal cumulative distribution and $\Phi$ is the standard normal probability density function. Using this EI function, the most promising candidate sample is selected by choosing $x^+$ that has the maximum EI value.
\begin{equation}
    x^{*} =\underset{x^+ \in DS }{\mathrm{argmax}} \; EI(x^+) 
\end{equation}
The newly selected sample $x^*$ is evaluated and is included in the evaluated data set, called $X$. Accordingly, the posterior probability distribution is estimated by the conditioning rules for Gaussian random variables, as below:   
\begin{gather}
    \mu^*= \mu(x^*)+\kappa(x^*,X)\kappa(X,X)^{-1}(f(X)-\mu(X))\\
    (\sigma^*)^2 = \kappa(x^*,x^*)-\kappa(x^*,X)\kappa(X,X)^{-1}\kappa(X,x^*)
\end{gather}
Constrained BO, which is an extension to standard BO meant to model infeasibility during the inequality-constrained optimization routine, is formulated and proposed by~\cite{gardner2014bayesian}. We use this formulation for our experimentation, and it models both function and constraint as Gaussian processes. Let $g$ be the constraint function that is unknown \textit{a priori}; the first step in this setting is to model $f$ and $g$ as Gaussian processes:    
\begin{gather}
    f\sim \mathcal{GP}(\mu_1(x),\kappa_1(x))\\
    g\sim \mathcal{GP}(\mu_2(x),\kappa_2(x))\\
    \mu_1 (x) = \mathbf{E}[f(x)]\\
    \kappa_1(x,x')=  \mathbf{E}[\{f(x)-\mu_1(x)\}\{f(x')-\mu_1(x')\}]\\
    \mu_2 (x) = \mathbf{E}[g(x)]\\
    \kappa_2(x,x')=  \mathbf{E}[\{g(x)-\mu_2(x)\}\{g(x')-\mu_2(x')\}]
\end{gather}
The improvement function in this case is modified as:
\begin{gather}
    I_C(x^+)=\Delta(x^+) max\{ 0,f(x^*)-f(x^+)\}\\
    \Delta(x^+) \in \{0,1\}
\end{gather}
 $\Delta(x^+)$ is a feasibility indicator function that is $1$ if $g(x^+)\leq 0$, and $0$ otherwise. It causes $\Delta(x^+)$ to be a Bernoulli random variable whose probability of getting a feasible design is: 
\begin{gather}
    PF(x^+):=Pr(g(x)\leq 0)= \int_{-\infty}^{0} P(g(x^+)|x^+,X) dg(x^+) 
\end{gather}
Due to the Gaussian behavior of $g(.)$, $\Delta(x^+)$ would be a univariate Gaussian random variable. The modified expected improvement to include the effect of infeasibility gives a joint acquisition function:  
\begin{eqnarray}
    EI_C(x^+) &=& \mathbf{E}[I_C(x^+)|x^+]\\
              &=& \mathbf{E}[{\Delta(x^+)I(x^+)|x^+}]\\
              &=& PF(x^+) EI(x^+)
\end{eqnarray}
 This joint acquisition function can be further optimized using standard optimization algorithms. Since our acquisition function has the property of being smooth and continuous, we used a two-step optimization to find $x^*$. The first step is Monte Carlo optimization and the second step is limited memory BFGS~\cite{fletcher2013practical} (see Figure \ref{fig:cbo}).      
    
\subsection{Integrated Toolchain}
\begin{figure*}[h!]
        \centering
        \captionsetup{justification=centering}
        \includegraphics[width=1.0\textwidth]{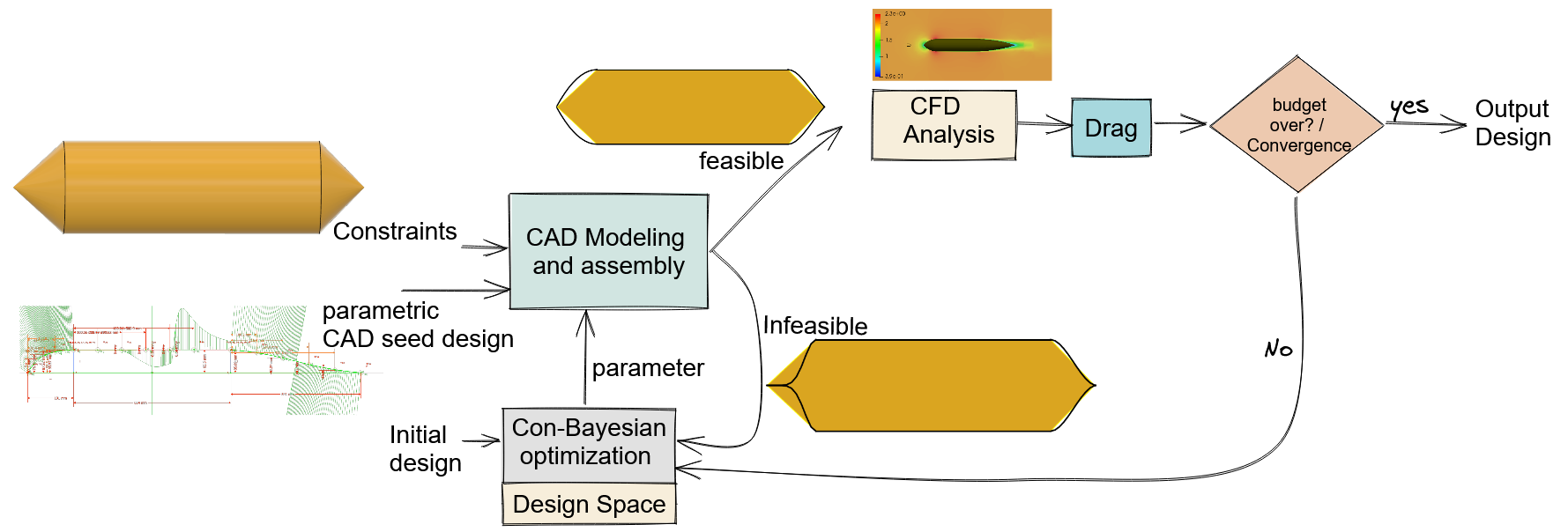}
        \caption{Optimization pipeline using integrated CAD and CFD tools with Bayesian Optimization}
        \label{fig:optimization pipeline}
    \end{figure*}

For design automation, exploration, and optimization, we integrated the necessary simulation tools for completely automated execution. To this end, we integrated the CAD design tool FreeCAD~\cite{riegel2016freecad} with the CFD simulation tool OpenFoam~\cite{jasak2007openfoam}. FreeCAD is used to design a parametric CAD model and generate the 3D CAD geometry from a given set of parameters along with its  stereolithography (STL) file without any manual intervention. OpenFoam uses this STL file to conduct fluid physics simulations via finite volume discretizations. Volume meshing is done using a castellated 3D parametric volumetric mesh which is further split and refined in the vicinity of the body surfaces. Other CFD simulation requirements, like solver settings and initial and boundary conditions, can also be set up from a Python environment. In the meshed volume, RANS with kw-SST turbulence fluid physics is solved, and the output of interest is fetched from OpenFoam and transferred back into the Python environment.  Accordingly, it is possible to both control and run the entire design optimization pipeline from a single Python environment. This integration of tools and capability to control the parameters and environmental conditions gives us the flexibility to run an optimization framework with design tools in the loop without human intervention (refer to Figure~\ref{fig:optimization pipeline}). 

\subsection{Parametric CAD Model and Baseline Packing Geometry}
For automatic design optimization, it was imperative to design a parametric CAD model with the flexibility and adaptability to be able to create a 3D model from a given set of parameters without manual intervention. The parametric CAD design should maintain the experimenter's assumptions in order to generate a \textit{valid} CAD design based on the given parameters. To ensure this, we use a stringent design methodology for completely constrained designs~\cite{hoffmann2001towards}. For the parametric CAD seed design, we used a Myring hull\cite{myring1976theoretical} as our baseline architecture. It is the most commonly used axisymmetrical hull shape, due to a number of advantages such as streamlined flow behavior and satisfactory geometry for both hydrodynamic and hydrostatic pressure. A Myring hull has three different body sections: nose, tail, and cylindrical center:
\begin{figure}[H]
        \centering
        \captionsetup{justification=centering}
        \includegraphics[width=0.45\textwidth]{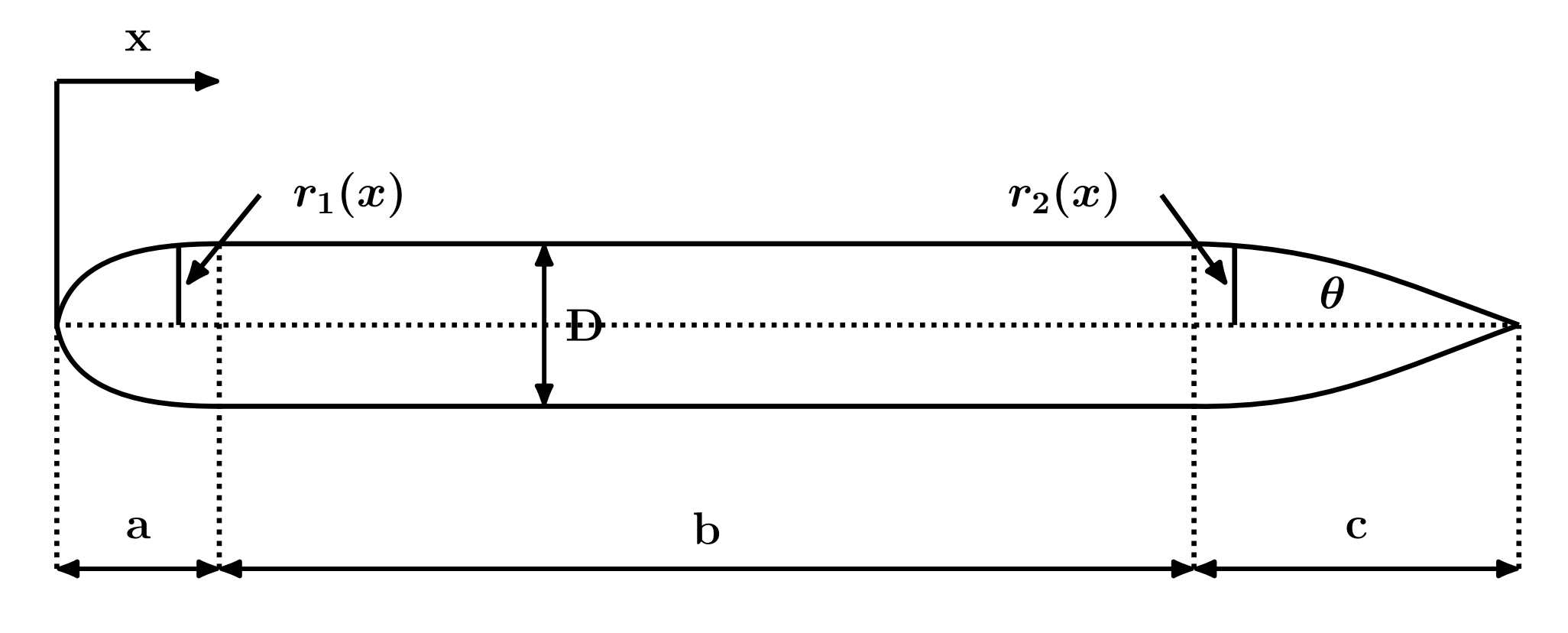}
        \caption{Myring hull: Geometry and parameters}
        \label{fig:myring}
\end{figure} 
The nose and tail equations for a Myring hull are given by:
\begin{gather}
    r_1(x) = \frac{1}{2}D\Big[1-\Big(\frac{x-a}{a}\Big)^2\Big]^{\frac{1}{n}} \\
    z= (x-a-b)\\
    r_2(x) = \frac{1}{2}D -\Big[ \frac{3D}{2c^2}-\frac{\tan \theta}{c}\Big]z^2+\Big[ \frac{D}{c^3}-\frac{\tan \theta}{c^2}\Big]z^3 .
\end{gather}
Here, $r_1$ and $r_2$ define the radius of the nose and tail of the hull at a distance, $x$, measured from the tip of the nose. Other body parameters are: $a$, $b$, $c$, $D$, $n$, and $\theta$, corresponding to nose length, body length, tail length, cylindrical body diameter, nose shaping parameter, and tail shaping parameter, respectively (see Figure \ref{fig:myring}).
    
\begin{figure}[H]
        \centering
        \captionsetup{justification=centering}
        \includegraphics[width=0.45\textwidth]{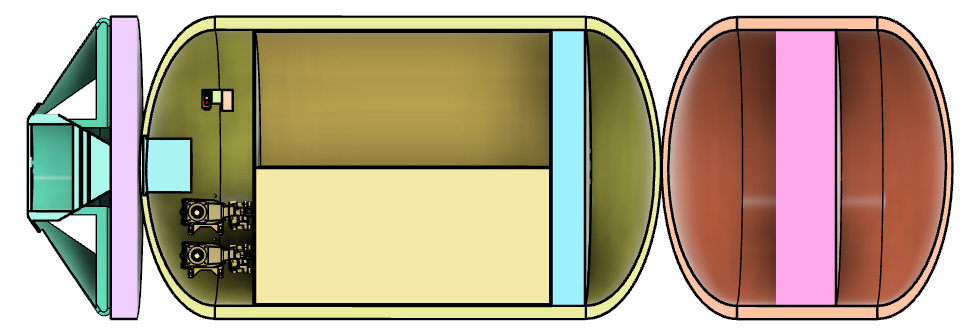}
        \caption{Selected components in the UUV in a specific packing configuration}
        \label{fig:cmp}
\end{figure}

\begin{figure}[H]
        \centering
        \captionsetup{justification=centering}
        \includegraphics[width=0.45\textwidth]{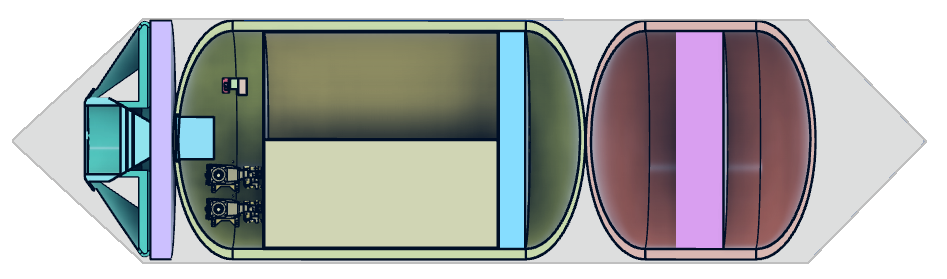}
        \vspace{-3pt}
        \caption{Selected components in baseline packed geometry}
        \label{fig:p_cmp}
\end{figure}

The baseline design used for internal component packing and placement comes from another automated tool~\cite{marotirapid}. Component selection and packing are not within the scope of this paper; however, three-dimensional packing of components in an arbitrary shape is an NP-complete problem. Based on the capabilities of our external component selection and packing tool, we utilized a simple design with conical end caps (nose and tail) and a cylindrical body to determine the exact required hull dimensions. These conical-shaped parametric designs are optimized to minimize internal hull volume while ensuring that packed components have no interferences. These baseline designs, however, are not optimal from the perspective of producing minimal-drag designs. Once components are packed and the parameters of a baseline design are found, this fixed geometry will act as a minimal constraint in the optimization of the hull design. (see Figures~\ref{fig:cmp} and \ref{fig:p_cmp}).

\subsection{Infeasible Design Heuristics}
Since a parametric Myring hull can assume a wide range of shapes \cite{myring1976theoretical}, the generated hull shape needs to be tested for interference with the baseline packed design. Any Myring hull parameters that cause interference with the baseline design are deemed to be infeasible. Since the computational cost of CAD assembly and running an interference test is much less than CFD simulation, the in-feasibility test on an optimized design is conducted during the CAD modeling and assembly stage (refer to Figure~\ref{fig:optimization pipeline}). Running full CFD analysis on an infeasible design is a waste of computational time and resources, and it delays the optimization process. To address this situation, we implemented a heuristic that works as follows (for a minimization problem): for an infeasible design that is detected during CAD assembly, return the maximum drag value to the optimizer for all evaluated samples up to that point, instead of running a full CFD analysis. The opposite can be done for a maximization problem. However, for starting the experiment we need at least one drag value of in-feasible design. Accordingly, we run the first infeasible design and store its drag value. 

\section{Design Experimentation and Results}
\label{sec:experiments}
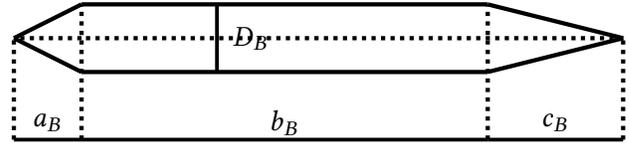
\begin{figure}[H]
    \centering
    \begin{tikzpicture}[scale=0.45,]
\centering

\tikzstyle{arrow} = [ultra thick, line width=0.5mm, align=center]

\coordinate (a) at(7,-30) {};

\coordinate (b) at(3,-30) {};
\coordinate (c) at(-9,-30) {};

\coordinate (e) at(3,-28) {};
\coordinate (f) at(-9,-28) {};

\coordinate (g) at(3,-26) {};
\coordinate (h) at(-9,-26) {};

\coordinate (i) at(7,-27) {};
\coordinate (j) at(-9,-27) {};

\coordinate (k) at(-9,-27) {};
\coordinate (l) at(-11,-27) {};
\coordinate (d) at(-11,-30) {};

\draw [arrow] (a) -- (b);
\draw [arrow] (b) -- (c);
\draw [arrow] (c) -- (d);

\draw [arrow] (-5,-28) -- (-5,-26);
\node[draw=none, ultra thick]at (-4, -27) {{\textbf{\Large{$D_B$}}}};

\draw [ultra thick] (e) -- (f) ;
\draw [ultra thick] (g) -- (h);
\draw [ultra thick](i) -- (e);
\draw [ultra thick](i) -- (g);
\draw [ultra thick](l) -- (f);
\draw [ultra thick](l) -- (h);

\draw [dotted, ultra thick](i) -- (j);

\draw [dotted, ultra thick](c) -- (h);
\draw [dotted, ultra thick](b) -- (g);
\draw [dotted, ultra thick](a) -- (i);
\draw [dotted, ultra thick](k) -- (l);
\draw [dotted, ultra thick](d) -- (l);

\node[draw=none, ultra thick] at (-10, -29.5){{\textbf{\Large{$a_B$}}}};
\node[draw=none, ultra thick] at (-3,-29.5) {\textbf{\Large{$b_B$}}};
\node[draw=none, ultra thick] at (5,-29.5) {\textbf{\Large{$c_B$}}};

\end{tikzpicture}
    \caption{Baseline 3D hull design with $a_{baseline}= 555$ cm, $b_{baseline}= 2664$ cm, $c_{baseline}= 512$ cm,  $D_{baseline}= 1026$ cm}
    \label{fig:baseline}
\end{figure}

In this section, we present two different experiments carried out using our optimization pipeline. In both cases, selected components are the same and consequently, the baseline packing geometry is identical. The operating conditions (i.e., the velocity of operation, initial and boundary conditions) and environmental conditions (e.g., turbulence intensity) are kept constant based on mission requirements. The baseline packing geometry is as shown in Figure~\ref{fig:baseline}. The design space ($DS$) of the search process is selected as shown in Table~\ref{tab:ds}.
  
\begin{table}
\centering
\normalsize
\captionsetup{justification=centering,format=plain,labelfont=bf}
\begin{tabular}{ccc} 
\toprule
\hfil \textbf{Symbol} & \hfil \textbf{ Minimum } & \hfil \textbf{ Maximum } \\
\midrule
 $a$ & $a_B$  & $a_B+2500$ mm \\
$c$ & $c_B$ & $c_B+2500$ mm \\
 $n$ & $0.1$ & $5.0$ \\
 $\theta$ & $0^o$ & $50^o$  \\
$l=a+b+c$ &  &  \\
\bottomrule
\end{tabular}
\vspace{0mm}
\caption{Range of design parameters for optimization}
\label{tab:ds}
\end{table}

\vspace{-10pt}
\subsection{Experiment 1}
\begin{figure}[h!]
        \centering
        \captionsetup{justification=centering}
        \includegraphics[width=0.5\textwidth]{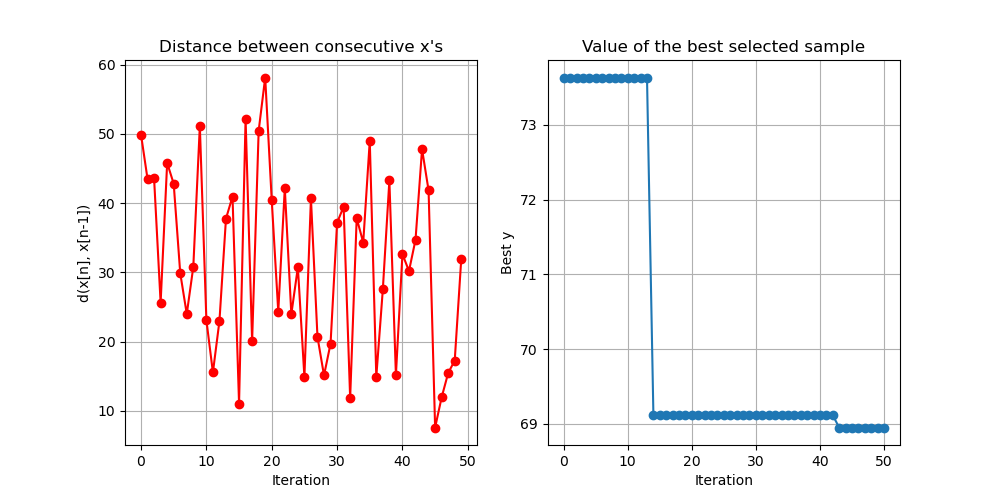}
        \caption{Optimization process vs number of evaluation/iteration: L2 distance between successive selected samples (left), Drag value of best-selected sample in Newton (right)  }
        \label{fig:opt_exp1}
\end{figure}

In this experiment, we only optimize the nose and tail shapes, defined by parameters $n$ and $\theta$. The range of the design space for optimization of parameters $n$ and $\theta$ is given in Table~\ref{tab:ds}. Due to it being a computationally costly process, we run $50$ iterations of optimization using our optimization pipeline. The most optimal design (shown in Figure~\ref{fig:opt_exp1}) has a drag value of approx $69$ Newtons. 
 \begin{figure}[h!]
        \centering
        \captionsetup{justification=centering}
        \includegraphics[width=0.5\textwidth]{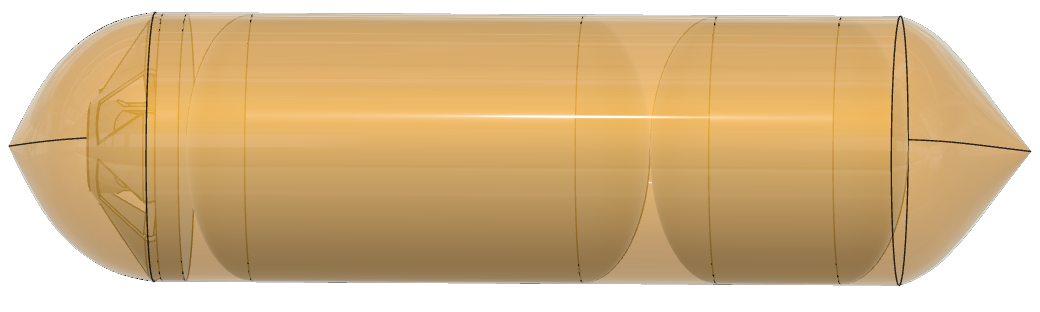}
        \caption{Optimal UUV hull shape with fixed nose and tail length. Optimal design parameters: $n= 1.0$; $\theta= 50.0$}
        \label{fig:res_exp1}
\end{figure} 

\subsection{Experiment 2}
In this experiment, we also optimized the nose and tail length (parameters $a$ and $b$) in addition to their shapes (parameters $n$ and $\theta$). The design space for optimization of all four variables is given in Table~\ref{tab:ds}. Again, we run 50 sequential optimization steps using our optimization pipeline. The most optimal design is shown in Figure~\ref{fig:opt_exp1} and had a drag value of approximately $36$ Newtons. This is a 50\% reduction in drag due to the streamlined nose and tail shapes and would save a large amount of energy consumption during real-world operation of the vehicle.

 \begin{figure}[h!]
        \centering
        \captionsetup{justification=centering}
        \includegraphics[width=0.45\textwidth]{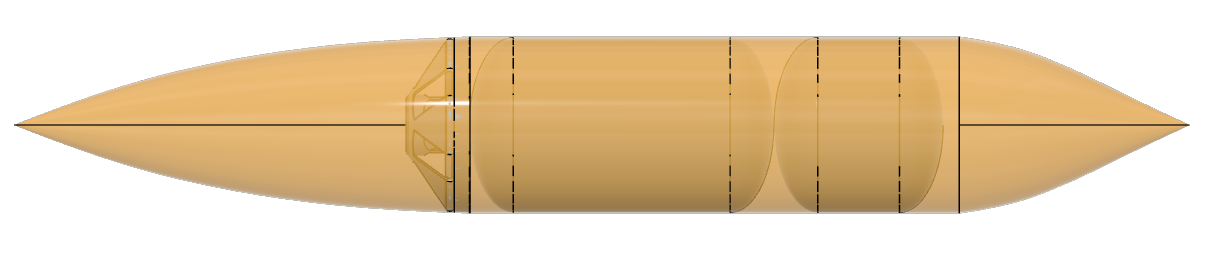}
        \caption{Optimal UUV hull shape with nose and tail length as a free parameter. Optimal design parameters: $a=2643.86$; $c=1348.72$; $n= 1.144$; $\theta= 22.03$}
        \label{fig:res_exp2}
\end{figure} 

 \begin{figure}[h!]
        \centering
        \captionsetup{justification=centering}
        \includegraphics[width=0.5\textwidth]{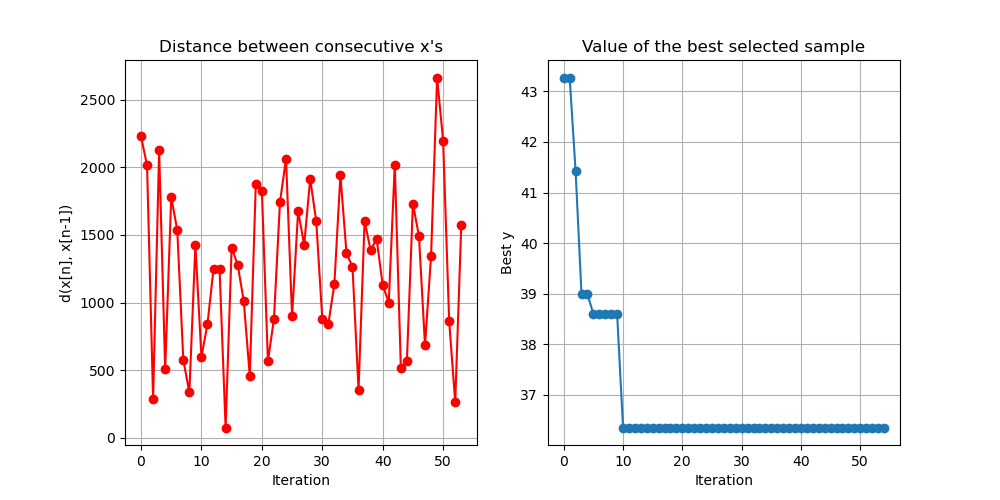}
        \caption{{Optimization process vs number of evaluation/iteration: L2 distance between successive selected samples (left), Drag value of the best-selected sample in Newton (right) }}
        \label{fig:opt_exp2}
\end{figure} 
\subsection{Analysis of result}
 In both experiments, the allocated budget was $50$ evaluations since the evaluation time was tens of minutes. But BO converges to optimal/near-optimal design in  a few iterations. In exp1 (refer to right side plot in figure \ref{fig:opt_exp1}) even in 12 iterations a near-optimal design was found and no significant further improvement is observed. In exp2 (refer to right plot in figure \ref{fig:opt_exp2}) only in 10 iterations the optimal design was found and no further improvement was observed.  This sample efficiency is due to the dynamic probabilistic modeling of the design space on labeled samples and state-of-the-art acquisition functions and accordingly costly optimization calculation. However, with the current multi-core implementation of BO, it takes milli-second to seconds for finding a new sample to evaluate and it is prudent to use BO in use cases where sample labeling and evaluation time can not be reduced beyond seconds.
\section{Related Work}
\label{sec:related_work}

One of the first well-known studies on optimizing UUV hull design for low drag was conducted by Gertler~\cite{gertler1950resistance} in $1950$. Later in 1976, Myring~\cite{myring1976theoretical} studied viscous-inviscid flow interaction methods to predict drag and concluded that there is low variability in body drag force when the nose or tail varies from slender to stout within a specific range, but it increases dramatically outside that range. To design shapes for better performance, bio-inspired hull shapes for UUVs are becoming popular~\cite{dong2020development}. To this end, Dong et al.~\cite{dong2020development} designed a gliding robotic fish with the streamlined shape of a whale shark. Stewart et al.~\cite{stewart2018design} designed a hybrid UAV-UUV system inspired by seabirds. A four-fin bio-inspired UUV was studied by Geder et al.~\cite{geder2013maneuvering}. They also showed that fish can achieve both high maneuverability and excellent gliding performance by equipping themselves with controllable fins and tails.

More recently, with extraordinary developments in computing capability and the maturity of mesh-based analysis tools, computational fluid dynamics (CFD) simulations are now widely applied to analyze UUV hydrodynamic performance. Most research is based on either the Reynolds-averaged Navier-Stokes (RANS) formulation or the large eddy simulation (LES). Since RANS treats viscous effects much better than potential flow theory and needs fewer computational resources, it is more frequently used than LES. With the advent of computer-aided design, traditional CFD can be leveraged inside an optimization loop to seek an optimal UUV design for given flow conditions; see, e.g., Alam et al.~\cite{alam2015design}. Many different optimization algorithms have been considered; for example, adjoint methods~\cite{jameson2003aerodynamic} and genetic algorithms~\cite{song2010research}.  \cite{vardhan2023fusion} integrated machine learning model with numerical simulation to get optimized design faster than the traditional methods. Schweyher et al.~\cite{schweyher1996optimization} used an evolutionary strategy- Genetic Algorithm to obtain a minimum-drag body. Application of Bayesian Optimization to find a minimum-drag shape was studied in 2D small arbitrary shapes by Eismann et al.~\cite{eismann2017shape} and an axisymmetric body of rotation by Vardhan et al \cite{vardhan2023search}. Both works did not consider constraint modeling during the design process. A deep neural network-based approach is used to study the effects of UUV shape on drag force by~\cite{vardhan2022data}. 

\section{Conclusion and Future works}
\label{sec:conclusion}
In this work, we developed an end-to-end design automation toolchain for constrained optimization problems for underwater vehicle hull design. We integrated the state-of-the-art AI-based optimization algorithm called Bayesian optimization along with the capability to handle constraints during the optimization process.  
Since this integrated tool is generic, the most interesting future work of interest is the extension and integration of other optimization algorithms with the current evaluation toolchain and comparing the performance of AI-based bayesian optimization with these optimization methods that are currently used in the domain of CFD-based optimization. To this end, it would be interesting to compare BO with other simulation-based optimization methods like GA,  Nelder Mead, and Particle Swarm Optimization along with the most used gradient-based optimization method i.e. adjoint-assisted optimization. A detailed comparative study can give us a better picture of the standing of AI-based optimizers in comparison to other existing optimization methods. 
\begin{acks}
This work is supported by DARPA through contract number FA8750-20-C-0537. Any opinions, findings, conclusions, or recommendations expressed are those of the authors and do not necessarily reflect the views of the sponsor.
\end{acks}

\balance
\bibliographystyle{unsrtnat}
\bibliography{References}

\begin{thebibliography}{30}
\providecommand{\natexlab}[1]{#1}
\providecommand{\url}[1]{\texttt{#1}}
\expandafter\ifx\csname urlstyle\endcsname\relax
  \providecommand{\doi}[1]{doi: #1}\else
  \providecommand{\doi}{doi: \begingroup \urlstyle{rm}\Url}\fi

\bibitem[Allard and Shahbazian(2014)]{allard2014unmanned}
Yannick Allard and Elisa Shahbazian.
\newblock Unmanned underwater vehicle (uuv) information study.
\newblock Technical report, OODA Technologies Inc Montreal, Quebec Canada,
  2014.

\bibitem[Vardhan et~al.(2021)Vardhan, Volgyesi, and
  Sztipanovits]{vardhan2021machine}
Harsh Vardhan, Peter Volgyesi, and Janos Sztipanovits.
\newblock Machine learning assisted propeller design.
\newblock In \emph{Proceedings of the ACM/IEEE 12th International Conference on
  Cyber-Physical Systems}, pages 227--228, 2021.

\bibitem[Alam et~al.(2015)Alam, Ray, and Anavatti]{alam2015design}
Khairul Alam, Tapabrata Ray, and Sreenatha~G Anavatti.
\newblock Design optimization of an unmanned underwater vehicle using low-and
  high-fidelity models.
\newblock \emph{IEEE Transactions on Systems, Man, and Cybernetics: Systems},
  47\penalty0 (11):\penalty0 2794--2808, 2015.

\bibitem[Vardhan and Sztipanovits(2022{\natexlab{a}})]{vardhan2022deep}
Harsh Vardhan and Janos Sztipanovits.
\newblock Deep learning based fea surrogate for sub-sea pressure vessel.
\newblock In \emph{2022 6th International Conference on Computer, Software and
  Modeling (ICCSM)}, pages 36--39. IEEE, 2022{\natexlab{a}}.

\bibitem[Vardhan and Sztipanovits(2021)]{vardhan2021rare}
Harsh Vardhan and Janos Sztipanovits.
\newblock Rare event failure test case generation in
  learning-enabled-controllers.
\newblock In \emph{2021 6th International Conference on Machine Learning
  Technologies}, pages 34--40, 2021.

\bibitem[Abbeel et~al.(2010)Abbeel, Coates, and Ng]{abbeel2010autonomous}
Pieter Abbeel, Adam Coates, and Andrew~Y Ng.
\newblock Autonomous helicopter aerobatics through apprenticeship learning.
\newblock \emph{The International Journal of Robotics Research}, 29\penalty0
  (13):\penalty0 1608--1639, 2010.

\bibitem[Vardhan and Sztipanovits(2022{\natexlab{b}})]{vardhan2022reduced}
Harsh Vardhan and Janos Sztipanovits.
\newblock Reduced robust random cut forest for out-of-distribution detection in
  machine learning models.
\newblock \emph{arXiv preprint arXiv:2206.09247}, 2022{\natexlab{b}}.

\bibitem[Clark(1961)]{clark1961greatest}
Charles~E Clark.
\newblock The greatest of a finite set of random variables.
\newblock \emph{Operations Research}, 9\penalty0 (2):\penalty0 145--162, 1961.

\bibitem[Mo{\v{c}}kus(1975)]{movckus1975bayesian}
Jonas Mo{\v{c}}kus.
\newblock On bayesian methods for seeking the extremum.
\newblock In \emph{Optimization techniques IFIP technical conference}, pages
  400--404. Springer, 1975.

\bibitem[Zhilinskas(1975)]{zhilinskas1975single}
AG~Zhilinskas.
\newblock Single-step bayesian search method for an extremum of functions of a
  single variable.
\newblock \emph{Cybernetics}, 11\penalty0 (1):\penalty0 160--166, 1975.

\bibitem[Gardner et~al.(2014)Gardner, Kusner, Xu, Weinberger, and
  Cunningham]{gardner2014bayesian}
Jacob~R Gardner, Matt~J Kusner, Zhixiang~Eddie Xu, Kilian~Q Weinberger, and
  John~P Cunningham.
\newblock Bayesian optimization with inequality constraints.
\newblock In \emph{ICML}, volume 2014, pages 937--945, 2014.

\bibitem[Rasmussen(2003)]{rasmussen2003gaussian}
Carl~Edward Rasmussen.
\newblock Gaussian processes in machine learning.
\newblock In \emph{Summer school on machine learning}, pages 63--71. Springer,
  2003.

\bibitem[Jones et~al.(1998)Jones, Schonlau, and Welch]{jones1998efficient}
Donald~R Jones, Matthias Schonlau, and William~J Welch.
\newblock Efficient global optimization of expensive black-box functions.
\newblock \emph{Journal of Global optimization}, 13\penalty0 (4):\penalty0
  455--492, 1998.

\bibitem[Fletcher(2013)]{fletcher2013practical}
Roger Fletcher.
\newblock \emph{Practical methods of optimization}.
\newblock John Wiley \& Sons, 2013.

\bibitem[Riegel et~al.(2016)Riegel, Mayer, and van Havre]{riegel2016freecad}
Juergen Riegel, Werner Mayer, and Yorik van Havre.
\newblock Freecad, 2016.

\bibitem[Jasak et~al.(2007)Jasak, Jemcov, Tukovic, et~al.]{jasak2007openfoam}
Hrvoje Jasak, Aleksandar Jemcov, Zeljko Tukovic, et~al.
\newblock Openfoam: A c++ library for complex physics simulations.
\newblock In \emph{International workshop on coupled methods in numerical
  dynamics}, volume 1000, pages 1--20. IUC Dubrovnik Croatia, 2007.

\bibitem[Hoffmann and Kim(2001)]{hoffmann2001towards}
Christoph~M Hoffmann and K-J Kim.
\newblock Towards valid parametric cad models.
\newblock \emph{Computer-Aided Design}, 33\penalty0 (1):\penalty0 81--90, 2001.

\bibitem[Myring(1976)]{myring1976theoretical}
DF~Myring.
\newblock A theoretical study of body drag in subcritical axisymmetric flow.
\newblock \emph{Aeronautical quarterly}, 27\penalty0 (3):\penalty0 186--194,
  1976.

\bibitem[Maroti et~al.()Maroti, Hedgecock, and Volgyesi]{marotirapid}
Miklos Maroti, Will Hedgecock, and Peter Volgyesi.
\newblock Rapid design space exploration with constraint programming.

\bibitem[Gertler(1950)]{gertler1950resistance}
Morton Gertler.
\newblock \emph{Resistance experiments on a systematic series of streamlined
  bodies of revolution: for application to the design of high-speed
  submarines}.
\newblock Navy Department, David W. Taylor Model Basin, 1950.

\bibitem[Dong et~al.(2020)Dong, Wu, Chen, Tan, and Yu]{dong2020development}
Huijie Dong, Zhengxing Wu, Di~Chen, Min Tan, and Junzhi Yu.
\newblock Development of a whale-shark-inspired gliding robotic fish with high
  maneuverability.
\newblock \emph{IEEE/ASME Transactions on Mechatronics}, 25\penalty0
  (6):\penalty0 2824--2834, 2020.

\bibitem[Stewart et~al.(2018)Stewart, Weisler, MacLeod, Powers, Defreitas,
  Gritter, Anderson, Peters, Gopalarathnam, and Bryant]{stewart2018design}
William Stewart, Warren Weisler, Marc MacLeod, Thomas Powers, Aaron Defreitas,
  Richard Gritter, Mark Anderson, Kara Peters, Ashok Gopalarathnam, and Matthew
  Bryant.
\newblock Design and demonstration of a seabird-inspired fixed-wing hybrid
  uav-uuv system.
\newblock \emph{Bioinspiration \& biomimetics}, 13\penalty0 (5):\penalty0
  056013, 2018.

\bibitem[Geder et~al.(2013)Geder, Ramamurti, Pruessner, and
  Palmisano]{geder2013maneuvering}
Jason~D Geder, Ravi Ramamurti, Marius Pruessner, and John Palmisano.
\newblock Maneuvering performance of a four-fin bio-inspired uuv.
\newblock In \emph{2013 OCEANS-San Diego}, pages 1--7. IEEE, 2013.

\bibitem[Jameson(2003)]{jameson2003aerodynamic}
Antony Jameson.
\newblock Aerodynamic shape optimization using the adjoint method.
\newblock \emph{Lectures at the Von Karman Institute, Brussels}, 2003.

\bibitem[Song et~al.(2010)Song, Zhu, and Liu]{song2010research}
Baowei Song, Qifeng Zhu, and Zhanyi Liu.
\newblock Research on multi-objective optimization design of the uuv shape
  based on numerical simulation.
\newblock In \emph{International Conference in Swarm Intelligence}, pages
  628--635. Springer, 2010.

\bibitem[Vardhan et~al.(2023)Vardhan, Volgyesi, and
  Sztipanovits]{vardhan2023fusion}
Harsh Vardhan, Peter Volgyesi, and Janos Sztipanovits.
\newblock Fusion of ml with numerical simulation for optimized propeller
  design.
\newblock \emph{arXiv preprint arXiv:2302.14740}, 2023.

\bibitem[Schweyher et~al.(1996)Schweyher, Lutz, and
  Wagner]{schweyher1996optimization}
H~Schweyher, Th~Lutz, and S~Wagner.
\newblock An optimization tool for axisymmetric bodies of minimum drag.
\newblock In \emph{2nd international airship conference,
  Stuttgart/Friedrichshafen}, pages 3--4, 1996.

\bibitem[Eismann et~al.(2017)Eismann, Bartzsch, and Ermon]{eismann2017shape}
Stephan Eismann, Stefan Bartzsch, and Stefano Ermon.
\newblock Shape optimization in laminar flow with a label-guided variational
  autoencoder.
\newblock \emph{arXiv preprint arXiv:1712.03599}, 2017.

\bibitem[Vardhan and Sztipanovits(2023)]{vardhan2023search}
Harsh Vardhan and Janos Sztipanovits.
\newblock Search for universal minimum drag resistance underwater vehicle hull
  using cfd.
\newblock \emph{arXiv preprint arXiv:2302.09441}, 2023.

\bibitem[Vardhan et~al.(2022)Vardhan, Timalsina, Volgyesi, and
  Sztipanovits]{vardhan2022data}
Harsh Vardhan, Umesh Timalsina, Peter Volgyesi, and Janos Sztipanovits.
\newblock Data efficient surrogate modeling for engineering design:
  Ensemble-free batch mode deep active learning for regression.
\newblock \emph{arXiv preprint arXiv:2211.10360}, 2022.

\end{thebibliography}
\end{document}